\documentclass[runningheads]{llncs}

\usepackage[final,year=2026]{eccv}

\usepackage{eccvabbrv}
\usepackage{graphicx}
\usepackage{booktabs}
\usepackage{tabularx}
\usepackage{comment}
\usepackage{makecell}
\usepackage{xcolor}
\usepackage{fvextra}
\usepackage{booktabs}
\usepackage{multirow}
\usepackage[accsupp]{axessibility}
\usepackage{hyperref}
\usepackage{orcidlink}

\DefineVerbatimEnvironment{PromptVerbatim}{Verbatim}{breaklines=true,breakanywhere=true,fontsize=\small}
\begin{document}

\title{CBX-Bench: A Human-Aligned MLLM Council for Benchmarking Concept Bottleneck Model Explanations}
\titlerunning{CBX-Bench}

\author{Yusuf Meric Karadag\inst{1}\orcidlink{0009-0008-3250-3928}
\and Gulay Oklan\inst{1}\orcidlink{0009-0006-6214-0465}
\and Seref Baris Cagliyan\inst{1}\orcidlink{0009-0003-5370-7915} 
\and Umut Ozdemir\inst{1}\orcidlink{0009-0007-7259-9007}
\and Emre Akbas\inst{1,2}\orcidlink{0000-0002-3760-6722}
}

\authorrunning{Y. Karadag et al.}

\institute{\inst{} Department of Computer Engineering, Middle East Technical University (METU)
\and \inst{} Robotics \& AI Center (ROMER), METU}

\maketitle

\begin{abstract}
Concept Bottleneck Models (CBMs) are designed to make visual classification interpretable by expressing predictions through human-understandable concepts.
Although interpretability is the central motivation for CBMs, they are still largely evaluated as predictive models by downstream classification accuracy, supplemented by isolated qualitative examples. 
This highlights a pressing need for quantitative measures, a challenge complicated by the infeasibility of ground-truth concept annotation at scale and the open nature of concept lists due to a lack of consensus.
To fill this gap, we develop a multimodal large language model  (MLLM) council that, given an image and its CBM explanation, produces an explanation quality score.
To ground and validate the council, we first conduct a human study to establish a ground-truth reference for CBM explanation quality: for an image, annotators compare explanations from two of LF-CBM, VLG-CBM, and CBM-Suite and choose the more useful one, or mark them as equally good or equally bad, yielding 2700 judgments over 900 image-comparison items on CUB-200, ImageNet-100, and Places365.
Against this human reference, our five-model council, consisting of open-weight MLLMs, recovers over 70\% of strict human preference rankings, rising to 83\% on items where human annotators unanimously agree.
 Building on this validated council, we introduce \textbf{CBX-Bench}, a public benchmark and leaderboard: authors of new CBMs can submit their model’s explanations, and CBX-Bench scores them with the council and maintains dataset-level rankings of explanation quality.
 CBX-Bench thus provides a human-aligned, scalable evaluation of CBM explanations beyond accuracy and isolated qualitative examples. The benchmark is available at \url{https://github.com/meric-karadag/cbx-bench}.


\keywords{Concept bottleneck models \and Explainable computer vision}
\end{abstract}

\section{Introduction}
\label{sec:introduction}

Concept Bottleneck Models (CBMs) are designed to make visual recognition interpretable by routing predictions through human-understandable concepts~\cite{koh2020concept}. Instead of mapping an image directly to a class label, a CBM first represents the image through intermediate concepts and then uses those concepts to make the final prediction. This structure is appealing because it promises explanations that expose what visual evidence the model is using, and because concept-level representations can in principle support inspection, comparison, and intervention. Recent CBM variants reduce the need for per-image concept annotations, construct concept vocabularies automatically, or use vision-language representations to build concept bottlenecks at scale~\cite{oikarinen2023label,srivastava2024vlg,tapli2026rethinking}.

\begin{figure}[t]
  \centering
  \includegraphics[width=\linewidth]{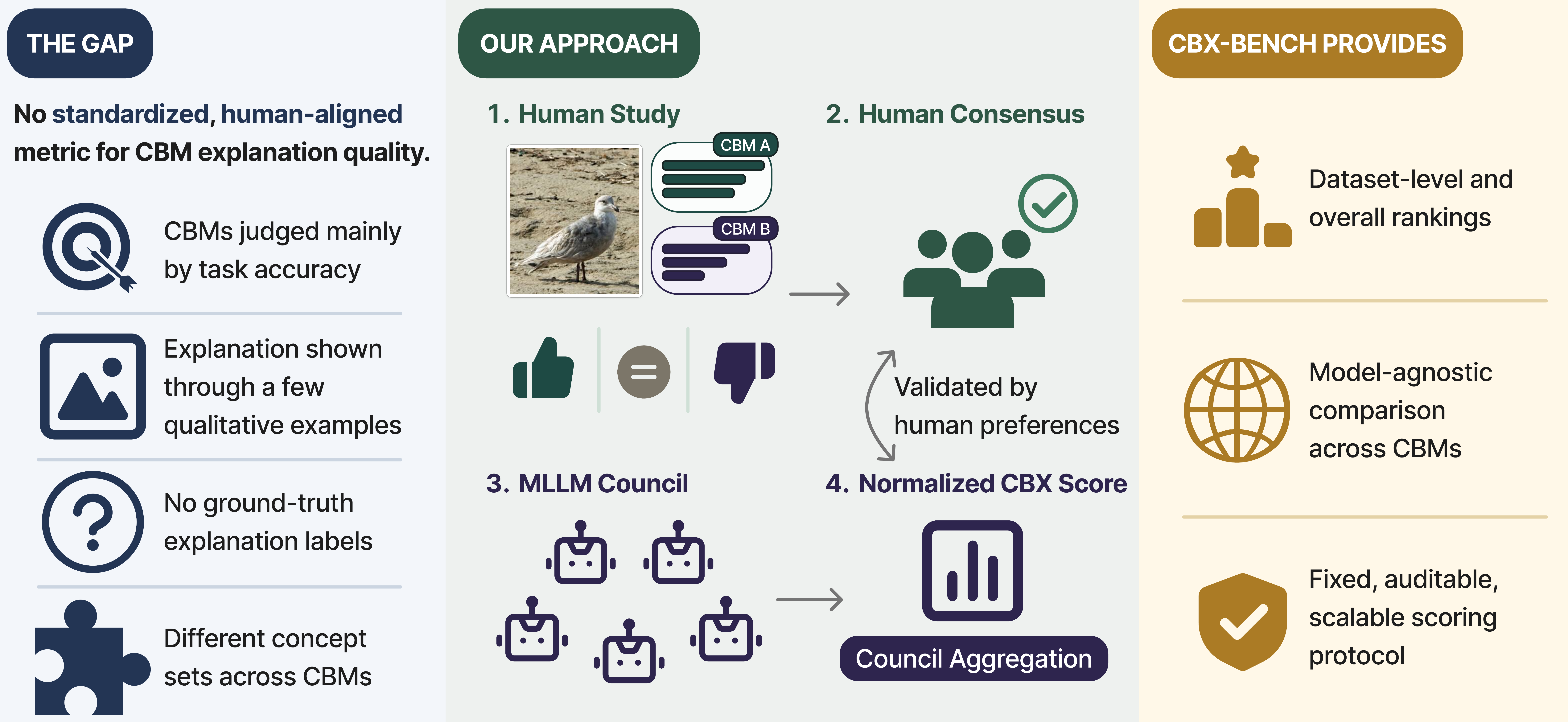}
\caption{CBX-Bench motivation and contribution. Existing CBM evaluations lack a
standardized, human-aligned metric for explanation quality. CBX-Bench addresses
this gap by validating MLLM judges with human preferences and using the resulting
council to produce normalized, model-agnostic CBM explanation rankings.}
  \label{fig:teaser}
\end{figure}

Despite this motivation, CBM explanations still lack a widely accepted, human-validated benchmark. Existing evaluations often emphasize downstream classification accuracy, concept prediction accuracy when annotations are available, or selected qualitative examples. These measurements are useful, but they do not directly quantify the quality of CBM explanations across the dataset. 
The problem is amplified by modern CBMs that use different concept sets. This makes direct concept-overlap metrics method-specific and prevents fair comparison across CBM families. 

A benchmark for CBM explanations should therefore compare the quality of  explanations themselves, without requiring shared concept vocabularies or ground-truth concept annotations. In this paper, 
we build such a reference through a human preference study. For a given image, human evaluators see two anonymized CBM explanations, the ground-truth class, and each model's prediction (see Appendix~\ref{app:user-study-ui}), then select which explanation better supports the image and the prediction, or mark the two explanations as equally good or equally bad. We run this study on subsets of CUB-200~\cite{wah2011cub}, ImageNet-100~\cite{deng2009imagenet}, and Places365~\cite{zhou2017places}; comparing LF-CBM~\cite{oikarinen2023label}, VLG-CBM~\cite{srivastava2024vlg}, and CBM-Suite~\cite{tapli2026rethinking} across 900 image-comparison items and 2700 human judgments. We collect three human responses for each image-and-two-CBMs triplet. 

To scale beyond human annotation, we evaluate multimodal large language models (MLLMs) as automated judges for CBM explanation quality. Our main protocol scores each CBM explanation independently on an integer scale. We then derive pairwise rankings from these scores and measure how well they recover the human preferences. This lets us select the MLLM judging strategy that is most aligned with human rankings. The best-performing method is a five-judge score-sum council using a 1--9 scoring scale, which recovers over 70\% of strict human preference rankings and  83\% on items with unanimous human agreement. We use this validated council to construct \textbf{CBX-Bench}, a standardized benchmark that assigns normalized explanation-quality scores and dataset-level rankings to CBM methods. \cref{fig:teaser} illustrates  the motivation and contributions of CBX-Bench. In addition to the human-study models, we evaluated CaBM~\cite{cagliyan2026caption} and V2C-CBM~\cite{he2025v2c} on CUB-200 and ImageNet-100 where their explanations are available.

CBX-Bench is intended as a living benchmark and a practical evaluation pathway for CBM explanation quality, where authors of new CBM methods can submit model predictions and concept explanations for standardized evaluation. The submitted explanations will be scored by the validated MLLM council, normalized to produce comparable CBX-Bench scores, and incorporated into dataset-level rankings.

Our contributions are:
\begin{itemize}
\item We introduce \textbf{CBX-Bench, the first benchmark for quantitative measurement of CBM explanation quality} without requiring concept ground truth or shared concept vocabularies.
\item We validate the benchmark protocol with a human preference study covering three CBMs, three datasets, 900 image-comparison items, and 2700 human judgments.
\item We develop a score-based MLLM judging protocol and identify a five-judge 1--9 score-sum council that best recovers human-derived CBM explanation rankings.
\item Using our benchmark, we evaluate recent, prominent CBMs  (LF-CBM, VLG-CBM, CBM-Suite, CaBM, and V2C-CBM) and provide  dataset-level rankings. Our benchmark is open to submissions to facilitate the quantitative evaluation of explanation qualities of future CBMs.
\end{itemize}

\section{Related Work}
\label{sec:related-work}

\paragraph{Concept bottleneck models.}
Concept Bottleneck Models (CBMs) route predictions through intermediate
human-understandable concepts, enabling users to inspect, edit, or intervene on
a model's reasoning~\cite{koh2020concept}. The original formulation learns
concept predictors from per-image concept annotations, which makes the approach
interpretable but difficult to scale to datasets where such annotations are not
available. Later work has therefore focused on reducing concept-supervision
requirements and constructing concept sets more automatically. Post-hoc
CBMs convert pretrained models into concept-based predictors by learning a
concept layer over existing representations, allowing concept-based explanations
without retraining the original model from scratch~\cite{yuksekgonul2022posthoc}.
LaBo uses language models to generate candidate concept descriptions and
vision-language representations to construct language-guided bottlenecks for
classification~\cite{yang2023language}. LF-CBM further removes the need for
manual concept annotations by constructing label-free bottlenecks from
vision-language representations~\cite{oikarinen2023label}. DN-CBM reverses the
usual define-then-learn pipeline by discovering concepts from model
representations and then assigning semantic names to the discovered concepts
before using them in the bottleneck layer~\cite{rao2024discover}. VLG-CBM uses
vision-language guidance and grounded visual evidence to improve the grounding
of learned concept bottlenecks~\cite{srivastava2024vlg}. More recent work
extends concept-based interpretability toward zero-shot and text-based settings:
EZPC explains CLIP zero-shot predictions through explicit concepts
~\cite{ozdemir2026ezpc}, while CaBM uses free-form image captions as a
bottleneck and derives concept-level explanations from the resulting text-based
classifier~\cite{cagliyan2026caption}. In parallel, CBM-Suite highlights
practical pitfalls in CBM training and evaluation and proposes improved
protocols for fairer CBM comparison~\cite{tapli2026rethinking}.

This expanding design space makes CBM explanations difficult to compare with a
single concept-level metric. Our work addresses this by evaluating the
explanations produced for the same image, rather than requiring all methods to
share the same concept vocabulary or annotation scheme.

\paragraph{Evaluation of CBMs.}
CBMs are commonly evaluated through a combination of predictive performance,
concept-level diagnostics, and qualitative examples. Task accuracy remains the
standard metric for assessing whether the bottleneck preserves classification
performance~\cite{koh2020concept,zarlenga2022concept,yuksekgonul2022posthoc,
yang2023language,oikarinen2023label}. When concept annotations are available,
concept accuracy or attribute-level metrics are also reported to measure
alignment between predicted concepts and human-defined labels
~\cite{koh2020concept,zarlenga2022concept}. As recent CBMs increasingly rely on
automatically generated or discovered concepts, evaluation has expanded toward
concept quality, grounding, sparsity, and faithfulness. For example, LaBo
selects language-generated concepts based on discriminative and diverse
information~\cite{yang2023language}; DN-CBM evaluates the semantic
meaningfulness of discovered and named concepts~\cite{rao2024discover};
VLG-CBM studies visual grounding and leakage through effective-concept metrics
such as NEC and ANEC~\cite{srivastava2024vlg}; and CBM-Suite analyzes
concept-set suitability, bottleneck bypassing, and backbone--VLM--concept-set
interactions~\cite{tapli2026rethinking}. Other work explicitly studies concept
leakage and its effect on bottleneck faithfulness~\cite{havasi2022addressing}.
Many CBM papers also include qualitative examples showing the top contributing
concepts for individual predictions.

While these evaluations capture important aspects of CBM behavior, they do not
directly compare the usefulness of instance-level explanations across methods.
CBX-Bench complements existing protocols by evaluating CBM explanations through
a human-preference validated, model-agnostic, score-based MLLM council.

\paragraph{Large language models as evaluators.}
LLM-as-a-judge protocols have become a practical way to scale evaluation for open-ended model outputs~\cite{li2025generation}.
MT-Bench and Chatbot Arena~\cite{zheng2023judging} showed that strong LLM judges can approximate human pairwise preferences for conversational responses, helping establish automated judging as a useful complement to human evaluation.
For CBM explanations, the evaluator must reason over both the image and the textual concept explanation.
MLLMs are therefore attractive judges for this setting, since they can inspect visual and textual inputs jointly.
At the same time, judge prompts are not neutral measurement instruments.
In pairwise evaluation, changing the order of candidate outputs can change the judge's preferred answer even when the candidate content is unchanged ~\cite{wang2024large,shi2025judging}.

Recent work has explored LLM juries as a way to improve the performance and robustness of LLM-as-a-judge evaluation.
PoLL replaces a single large LLM judge with a panel of smaller judges drawn from different model families, reporting stronger correlation with human judgments and reduced intra-model bias in LLM generation evaluation~\cite{verga2024replacing}.
Motivated by this, CBX-Bench uses MLLM judges from different model families and aggregates their answers to find the best human-aligned protocol as our benchmark scorer.

\section{Methodology}
\label{sec:methodology}

\begin{figure}[t]
  \centering
  \includegraphics[width=\linewidth]{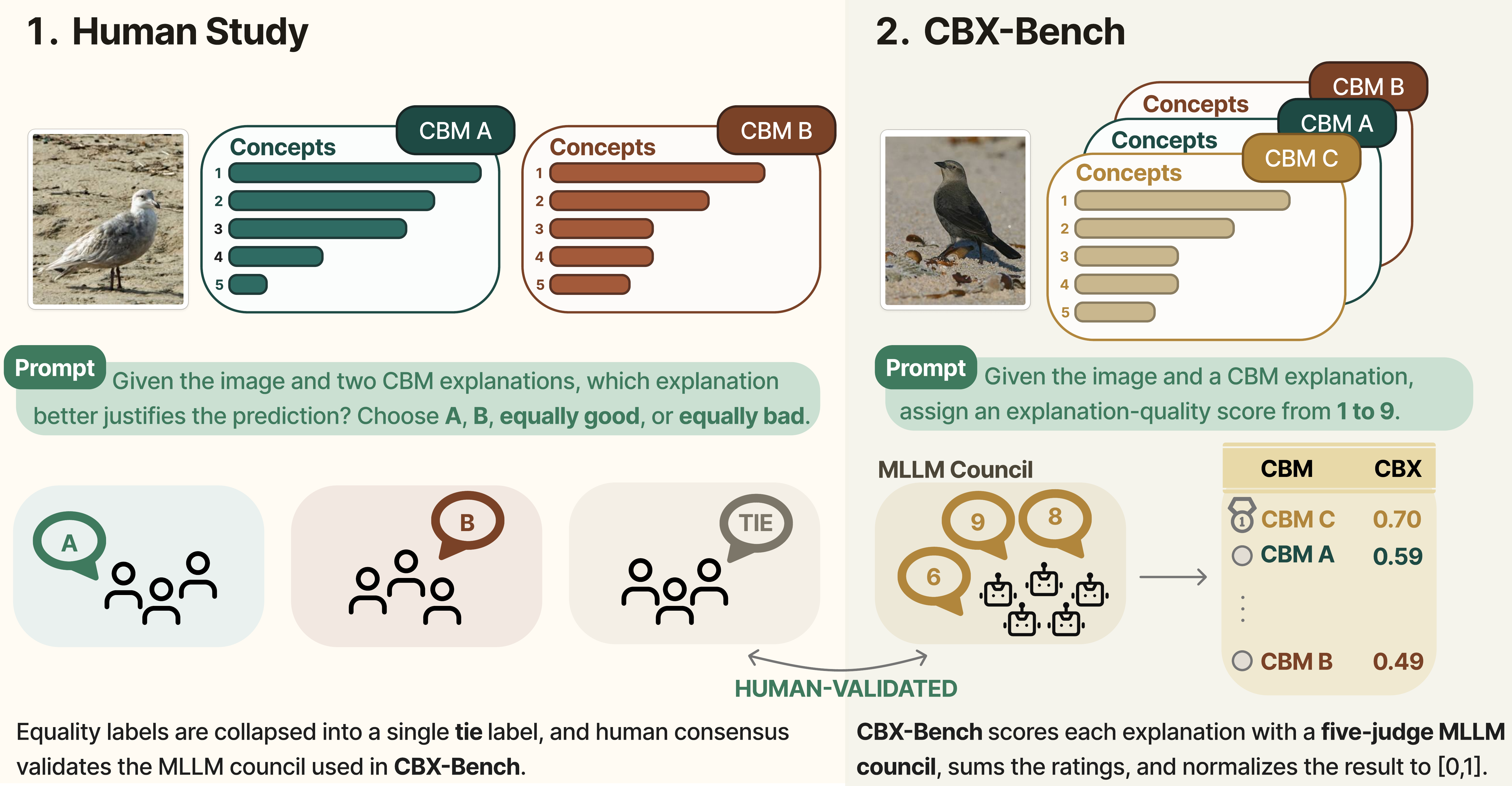}
\caption{CBX-Bench methodology. The human study collects pairwise preferences
over CBM explanations for the same image, collapses equality labels into a tie
for agreement analysis, and uses human consensus to validate score-based MLLM
judges. The selected score-sum council then scores each explanation
independently and normalizes the aggregate score to define CBX-Bench rankings.}
  \label{fig:methodology}
\end{figure}

\subsection{Evaluation Protocol and Human Study}
\label{sec:protocol}
\label{sec:user-study}
Figure~\ref{fig:methodology} summarizes our methodology. First, we conduct a human study to validate the evaluation protocol; then, the selected MLLM council scales the protocol to CBX-Bench scoring.

We evaluate CBM explanation quality through pairwise human comparison. Each item shows an image, its ground-truth class, two CBM predictions, and the two ranked concept explanations. Annotators choose one of four labels: prefer CBM A, prefer CBM B, \emph{equally good}, or \emph{equally bad}. The equality labels let annotators distinguish positive and negative ties while avoiding the harder task of calibrating absolute human scores across datasets, classes, and model families. For MLLM--human alignment, we merge \emph{equally good} and \emph{equally bad} into a single tie label.

For each CBM prediction, we display the top 5 concepts for the predicted class. Concept contribution is computed as the concept activation multiplied by the corresponding linear classifier weight. We rank concepts by this contribution, retain the top five, normalize their displayed contributions to sum to one, and omit concepts with normalized contribution below $0.01$ to remove numerical artifacts from near-zero activations.

We use pairwise comparison because it is better suited to this setting than absolute human scoring. Asking annotators to calibrate numerical explanation-quality scores across datasets, classes, and model families would introduce an additional source of subjectivity, whereas pairwise judgments directly match our goal of comparing CBM explanations.

This protocol is model-agnostic. Two CBMs may use different concept vocabularies, but their explanations can still be compared for the same image and decision. It also defines the reference structure for automated evaluation: independent MLLM scores can be judged by how well their induced pairwise relations recover human preferences.

The human study compares LF-CBM~\cite{oikarinen2023label}, VLG-CBM~\cite{srivastava2024vlg}, and CBM-Suite~\cite{tapli2026rethinking} on CUB-200~\cite{wah2011cub}, ImageNet-100 derived from ImageNet~\cite{deng2009imagenet}, and Places365~\cite{zhou2017places}. As shown in Table~\ref{tab:study-design}, each dataset contributes 
100 images and three CBM pairs, resulting in 900 image-comparison items 
and 2,700 human annotations in total.

\begin{table}[t]
\centering
\caption{Human study design. Each dataset contributes 100 images and three pairwise CBM comparisons per image.}
\label{tab:study-design}

\renewcommand{\arraystretch}{1.15}
\setlength{\tabcolsep}{8pt}

\begin{tabular}{lrrrr}
\toprule
Dataset & Images & CBM pairs & Items & Human votes \\
\midrule
CUB-200      & 100 & 3 & 300 & 900 \\
ImageNet-100 & 100 & 3 & 300 & 900 \\
Places365   & 100 & 3 & 300 & 900 \\
\addlinespace[2pt]
\midrule
\textbf{Total} & \textbf{300} & \textbf{3} & \textbf{900} & \textbf{2700} \\
\bottomrule
\end{tabular}
\end{table}

To avoid cherry-picking evaluation images, we construct each 100-image dataset subset by first randomly selecting 100 distinct classes and then sampling one image per class. Since CBM explanations are generally better for correctly classified examples, we search over random seeds and retain the first subset for which each human-study CBM's accuracy is within two percentage points of its full test-set accuracy. This keeps the sampled subsets diverse while avoiding a sample that accidentally favors or penalizes a particular CBM. After subsampling, all CBMs are evaluated using the same 100 images per dataset which includes correctly and incorrectly classified examples.

Each item is annotated by three participants who were primarily undergraduate and graduate students, using the web interface shown in Appendix~\ref{app:user-study-ui}. After collapsing the two equality labels into tie, we assign a consensus label when annotators agree unanimously $(3/3)$ or by majority $(2/3)$; items without such agreement are marked as no consensus and excluded from MLLM--human alignment. The protocol intentionally evaluates explanation preference rather than classification accuracy because a correct prediction can still receive a weak explanation judgment if its concepts are generic, spurious, or poorly grounded in the image.

\subsection{Multimodal Large Language Model (MLLM) Judges}
\label{sec:vlm-judges}

We evaluate MLLMs as scalable judges for CBM explanation quality. Each judge 
scores a single CBM explanation given the image, the ground-truth class, the 
CBM prediction, and the ranked concepts supporting that prediction. We use five popular open-weight MLLMs: InternVL3.5-8B~\cite{wang2025internvl3_5}, Qwen3.5-4B, Qwen3.5-9B~\cite{qwen35blog}, Gemma4-E2B, and Gemma4-E4B~\cite{gemma4_2026}. We use MLLMs in a zero-shot fashion: no judge is trained, fine-tuned, or given human preference labels at inference time. Human annotations are used only after inference to compare candidate judging protocols and find the most human-aligned MLLM council.

\paragraph{Independent score-based judging.}
Our main MLLM protocol scores each CBM explanation independently on an integer scale. We evaluate 1--5, 1--7, and 1--9 scales.
For a pair of CBM explanations, scores induce a pairwise label: the higher-scored explanation is preferred, and equal scores produce a tie. For an image with three CBMs, the three induced pairwise relations define an instance-level ranking over the displayed explanations. We compare these MLLM-derived relations with the corresponding
human consensus labels and compute, for each image, the fraction of
human-consensus pairwise relations recovered by the MLLM ranking. For
reproducibility all MLLM evaluations use greedy decoding. Prompt templates for evaluation are given in Appendix~\ref{app:vlm-prompts}.

\paragraph{Council aggregation.}
We aggregate the five MLLM judges in two ways. A vote-majority council first converts each judge's score pair into a pairwise label, then votes over the five labels, returning a tie when no label reaches a majority. A score-sum council instead sums the five scores assigned to each CBM explanation and derives pairwise labels from the aggregate scores. We use the score-sum council as the scoring mechanism for CBX-Bench after
selecting it through the human-alignment analysis reported in Sec.~\ref{sec:results}.

\subsection{CBX-Bench Scoring and Rankings}
\label{sec:cbx-bench}

CBX-Bench uses the selected score-sum council as a fixed scoring protocol for
CBM explanation quality. 
For image $i$, method $m$, and judge $j$,
let $s_{i,m,j}\in\{1,\ldots,9\}$ be the assigned explanation-quality score. We
sum the five judge scores and normalize the aggregate score from $[5,45]$ to $[0,1]$:
\[
q_{i,m}=\frac{\sum_{j=1}^{5}s_{i,m,j}-5}{40}.
\]
For each dataset $D$, the CBX-Bench score of method $m$ is the mean normalized
score over the evaluated images,
\[
Q_{D,m}=\frac{1}{|D|}\sum_{i\in D} q_{i,m}.
\]
The resulting CBX-Bench scores define a fixed, auditable, and human-aligned
evaluation signal for both existing and future CBMs. For a new submission, authors provide predictions and concept contribution scores
for the fixed evaluation images. CBX-Bench scores these submitted explanations
using the validated MLLM council. Scores are reported per dataset and overall,
enabling a standardized ranking of CBMs by explanation quality. Because every
method is evaluated on the same image set and with the same scoring protocol,
CBX-Bench supports fair comparison across CBMs with different architectures,
concept vocabularies, and explanation formats.

\section{Results}
\label{sec:results}

\paragraph{CBM explanation comparison is difficult even for humans.}
\begin{table}[t]
\centering
\caption{Human consensus strength across 900 pairwise CBM explanation comparisons in our human-study.}
\label{tab:human-consensus-strength}
\small
\renewcommand{\arraystretch}{1}
\setlength{\tabcolsep}{6pt}

\begin{tabular}{lrr}
\toprule
Consensus strength & Items & Outcome breakdown \\
\midrule
No consensus        & 109 (12.1\%) & -- \\
Majority consensus $(2/3)$ & 508 (56.4\%) & 380 CBM preferences \& 128 ties \\
Unanimous consensus $(3/3)$ & 283 (31.4\%) & 239 CBM preferences \& 44 ties \\
\addlinespace[2pt]
\midrule
{Total} & {900 (100.0\%)} & {619 CBM preferences \& 172 ties} \\
\bottomrule
\end{tabular}
\end{table}

Annotators reach consensus on 791/900 image-comparison items, but only 283 items (31.4\%) have unanimous 3/3 agreement (Table~\ref{tab:human-consensus-strength}). Across all 900 items, 619 (68.8\%) yield a clear preference for one CBM, 172 (19.1\%) yield a tie consensus, and 109 (12.1\%) do not yield consensus. Thus, pairwise explanation comparison is already a difficult judgment task for humans. This supports our decision to collect pairwise human preferences rather than absolute human scores: pairwise choices reduce annotator calibration burden, while still exposing disagreement and equality when the comparison is genuinely ambiguous or hard.

\paragraph{Score-based MLLMs recover instance-wise human rankings.}

\begin{table*}[t]
    \centering
    \caption{Instance-wise ranking recovery on CBM-only human consensus items. All human pairwise relations are strict preferences. Values are percentages; bold marks the best result within each scale and column.}
    \label{tab:instance-rank-alignment-scoring-cbm-only}
    \small
    \setlength{\tabcolsep}{1pt}
    \begin{tabular}{llc|c|c|c}
    \toprule
    Scale & Predictor & CUB200 Align. & IN100 Align. & Places365 Align. & Overall Align. \\
    \midrule
    \multirow{7}{*}{\rotatebox[origin=c]{90}{1--5}}
     & InternVL3.5-8B & 37.0 & 37.5 & 48.4 & 40.7 \\
     & Qwen3.5-4B & 56.5 & 51.0 & 52.7 & 53.5 \\
     & Qwen3.5-9B & 49.1 & 49.0 & 60.2 & 52.5 \\
     & Gemma4-E2B & 38.0 & 41.7 & 38.7 & 39.4 \\
     & Gemma4-E4B & 48.1 & 53.1 & 46.2 & 49.2 \\
     & Vote-majority council & 43.5 & 49.0 & 53.8 & 48.5 \\
     & Score-sum council & \textbf{66.7} & \textbf{61.5} & \textbf{74.2} & \textbf{67.3} \\
    \midrule
    \multirow{7}{*}{\rotatebox[origin=c]{90}{1--7}}
     & InternVL3.5-8B & 55.6 & 49.0 & 55.9 & 53.5 \\
     & Qwen3.5-4B & 57.4 & 52.1 & 57.0 & 55.6 \\
     & Qwen3.5-9B & 56.5 & 53.1 & 63.4 & 57.6 \\
     & Gemma4-E2B & 41.7 & 45.8 & 36.6 & 41.4 \\
     & Gemma4-E4B & 44.4 & 45.8 & 36.6 & 42.4 \\
     & Vote-majority council & 48.1 & 53.1 & 50.5 & 50.5 \\
     & Score-sum council & \textbf{67.6} & \textbf{63.5} & \textbf{71.0} & \textbf{67.3} \\
    \midrule
    \multirow{7}{*}{\rotatebox[origin=c]{90}{1--9}}
     & InternVL3.5-8B & 44.4 & 47.9 & 50.5 & 47.5 \\
     & Qwen3.5-4B & 53.7 & 52.1 & 55.9 & 53.9 \\
     & Qwen3.5-9B & 65.7 & 60.4 & 63.4 & 63.3 \\
     & Gemma4-E2B & 47.2 & 46.9 & 41.9 & 45.5 \\
     & Gemma4-E4B & 49.1 & 54.2 & 41.9 & 48.5 \\
     & Vote-majority council & 48.1 & 53.1 & 53.8 & 51.5 \\
     & Score-sum council & \textbf{68.5} & \textbf{68.8} & \textbf{75.3} & \textbf{70.7} \\
    \bottomrule
    \end{tabular}
    \end{table*}

\begin{table*}[t]
\centering
\caption{Instance-wise ranking recovery on human consensus items including ties. Values are percentages of recovered pairwise relations per image; bold marks the best result within each scale and column.}
\label{tab:instance-rank-alignment-scoring-ties}
\small
    \setlength{\tabcolsep}{1    pt}
    \begin{tabular}{llc|c|c|c}
\toprule
Scale & Predictor & CUB200 Align. & IN100 Align. & Places365 Align. & Overall Align. \\
\midrule
\multirow{7}{*}{\rotatebox[origin=c]{90}{1--5}}
 & InternVL3.5-8B & 41.4 & 44.6 & 50.5 & 45.6 \\
 & Qwen3.5-4B & 56.1 & 45.1 & 54.3 & 51.9 \\
 & Qwen3.5-9B & 48.5 & 49.2 & 57.6 & 51.9 \\
 & Gemma4-E2B & 43.9 & 42.1 & 42.4 & 42.8 \\
 & Gemma4-E4B & 51.5 & 50.8 & 50.0 & 50.7 \\
 & Vote-majority council & 51.5 & 50.8 & \textbf{59.5} & 54.1 \\
 & Score-sum council & \textbf{57.1} & \textbf{51.3} & 56.7 & \textbf{55.1} \\
\midrule
\multirow{7}{*}{\rotatebox[origin=c]{90}{1--7}}
 & InternVL3.5-8B & 55.6 & 49.2 & 54.8 & 53.2 \\
 & Qwen3.5-4B & 51.0 & 48.2 & 54.8 & 51.4 \\
 & Qwen3.5-9B & 55.6 & \textbf{53.3} & 57.6 & \textbf{55.6} \\
 & Gemma4-E2B & 46.0 & 39.0 & 38.1 & 41.0 \\
 & Gemma4-E4B & 47.5 & 48.2 & 43.8 & 46.4 \\
 & Vote-majority council & 52.5 & \textbf{53.3} & \textbf{58.1} & 54.7 \\
 & Score-sum council & \textbf{56.6} & 51.3 & 56.7 & 54.9 \\
\midrule
\multirow{7}{*}{\rotatebox[origin=c]{90}{1--9}}
 & InternVL3.5-8B & 47.0 & 50.8 & 52.4 & 50.1 \\
 & Qwen3.5-4B & 50.0 & 49.7 & 53.8 & 51.2 \\
 & Qwen3.5-9B & \textbf{59.1} & \textbf{57.4} & 56.2 & \textbf{57.5} \\
 & Gemma4-E2B & 46.0 & 46.2 & 41.0 & 44.3 \\
 & Gemma4-E4B & 46.5 & 51.3 & 42.9 & 46.8 \\
 & Vote-majority council & 46.0 & 53.3 & 54.8 & 51.4 \\
 & Score-sum council & 58.6 & 54.9 & \textbf{57.1} & 56.9 \\
\bottomrule
\end{tabular}
\end{table*}

For $N$ CBMs, human preferences and MLLM-derived score rankings each induce
$\binom{N}{2}$ pairwise relations.
We define human--MLLM alignment as the instance-wise ranking recovery which is the fraction of these relations that agree between
the human and MLLM rankings.
In our human-study $N=3$: for each image, the three pairwise human
labels define an ordering over the three CBM explanations whenever the relevant
comparisons have consensus, and the MLLM scores define the corresponding three pairwise relations.

Table~\ref{tab:instance-rank-alignment-scoring-cbm-only} reports the strict CBM-only setting, where all human relations are CBM preferences, excluding tie cases. A uniform random strict ranking over three CBMs has expected pairwise recovery of 50.0\%. The best method, the 1--9 score-sum council, reaches 70.7\% overall, with 68.5\% on CUB-200, 68.8\% on ImageNet-100, and 75.3\% on Places365. This makes the 1--9 score-sum council our primary MLLM evaluation method.

Table~\ref{tab:instance-rank-alignment-scoring-ties} includes human ties. With three CBMs, there are 13 possible rankings: six strict rankings, six rankings with one tied pair, and one all-tie ranking. Under a uniform weak-ranking chance reference, a pairwise relation matches with probability $59/169=34.9\%$. The 1--9 score-sum council reaches 56.9\% overall, while the best individual judge reaches 57.5\%. Thus, score-based MLLM evaluation remains well above chance even when equality judgments are included, although ties continue to be harder than clear preferences.

\begin{table}[t]
\centering
\caption{Instance-wise ranking recovery for the 1--9 score-sum council on images where human annotators are unanimous (3/3 agree) on the relevant pairwise comparisons.}
\label{tab:unanimous-rank-alignment-scoring}
\small
\setlength{\tabcolsep}{5pt}
\begin{tabular}{lrrrr}
\toprule
Unanimous Human Preference & CUB200 & ImageNet100 & Places365 & Overall \\
\midrule
a CBM & 66.7 & 88.9 & 88.9 & 83.3 \\
a CBM or tie  & 66.7 & 76.2 & 80.0 & 75.6 \\

\bottomrule
\end{tabular}
\end{table}

The alignment improves when the human reference is stronger. On images where the relevant human annotations are unanimous, the 1--9 score-sum council reaches 75.6\% ranking recovery when ties are included and 83.3\% in the CBM-only setting (Table~\ref{tab:unanimous-rank-alignment-scoring}). These results show that MLLM--human alignment increases with human consensus strength.

\paragraph{CBX-Bench produces dataset-level CBM explanation rankings.}

\providecommand{\kaynak}[1]{\hfill{\scriptsize\textcolor{gray}{#1}}}
\begin{table*}[t]
    \centering
    \caption{CBX-Bench explanation-quality scores. Scores are computed with the 1--9 score-sum MLLM council and normalized from the five-judge sum
    range $[5,45]$ to $[0,1]$. Bold marks the highest CBX-Bench score for each
    dataset. Test accuracy is the corresponding full-test-set classification accuracy}
    \label{tab:benchmark-scores-scoring}

    \renewcommand{\arraystretch}{1.18}
    \setlength{\tabcolsep}{6pt}
    \footnotesize

    \begin{tabular}{@{}
    >{\centering\arraybackslash}m{0.10\textwidth}
    >{\raggedright\arraybackslash}m{0.36\textwidth}
    >{\centering\arraybackslash}m{0.24\textwidth}
    >{\centering\arraybackslash}m{0.16\textwidth}
    @{}}
    \toprule
    Dataset
    & CBX-Bench Ranking
    & CBX-Bench Score 
    & Test acc. (\%) \\
    \midrule
    
    \phantom{\rotatebox[origin=c]{90}{CUB200}}%
    \llap{\smash{\raisebox{-1.5ex}{\rotatebox[origin=c]{90}{CUB200}}}}
    &
    \begin{tabular}[c]{@{}p{0.24\textwidth}@{\hspace{0pt}}p{0.16\textwidth}@{}}
    1. CaBM & \kaynak{(ECCV'26)} \\
    2. VLG-CBM & \kaynak{(NeurIPS'24)} \\
    3. CBM-Suite & \kaynak{(CVPR'26)} \\
    4. LF-CBM & \kaynak{(ICLR'23)} \\
    5. V2C-CBM & \kaynak{(AAAI'25)}
    \end{tabular}
    &
    \begin{tabular}[c]{@{}c@{}}
    \textbf{0.7088} \\
    0.5902 \\
    0.4905 \\
    0.4855 \\
    0.3088
    \end{tabular}
    &
    \begin{tabular}[c]{@{}c@{}}
    76.92 \\
    75.79 \\
    \textbf{86.73} \\
    74.31 \\
    83.00
    \end{tabular}
    \\
    \midrule

    \phantom{\rotatebox[origin=c]{90}{ImageNet100}}%
    \llap{\smash{\raisebox{-0.5ex}{\rotatebox[origin=c]{90}{ImageNet100}}}}
    &
    \begin{tabular}[c]{@{}p{0.24\textwidth}@{\hspace{0pt}}p{0.16\textwidth}@{}}
    1. CaBM & \kaynak{(ECCV'26)} \\
    2. VLG-CBM & \kaynak{(NeurIPS'24)} \\
    3. CBM-Suite & \kaynak{(CVPR'26)} \\
    4. LF-CBM & \kaynak{(ICLR'23)} \\
    5. V2C-CBM & \kaynak{(AAAI'25)}
    \end{tabular}
    &
    \begin{tabular}[c]{@{}c@{}}
    \textbf{0.7372} \\
    0.5528 \\
    0.5495 \\
    0.5000 \\
    0.3706
    \end{tabular}
    &
    \begin{tabular}[c]{@{}c@{}}
    75.12 \\
    73.15 \\
    81.56 \\
    71.95 \\
    \textbf{84.10}
    \end{tabular}
    \\
    \midrule

    \rotatebox[origin=c]{90}{Places365}
    &
    \begin{tabular}[c]{@{}p{0.24\textwidth}@{\hspace{0pt}}p{0.16\textwidth}@{}}
    1. CBM-Suite & \kaynak{(CVPR'26)} \\
    2. LF-CBM & \kaynak{(ICLR'23)} \\
    3. VLG-CBM & \kaynak{(NeurIPS'24)}
    \end{tabular}
    &
    \begin{tabular}[c]{@{}c@{}}
    \textbf{0.6047} \\
    0.5955 \\
    0.5817
    \end{tabular}
    &
    \begin{tabular}[c]{@{}c@{}}
    \textbf{54.64} \\
    43.68 \\
    41.92
    \end{tabular}
    \\

    \bottomrule
    \end{tabular}
\end{table*}

Having validated the 1--9 score-sum council against human preferences, we apply
it as the CBX-Bench scoring protocol to all CBM explanations in the 100-image
sample from each dataset. Table~\ref{tab:benchmark-scores-scoring} reports the
resulting normalized explanation-quality scores and dataset-level rankings. On
CUB-200 and ImageNet-100, where CaBM~\cite{cagliyan2026caption} and
V2C-CBM~\cite{he2025v2c} are also available, CaBM obtains the highest
CBX-Bench score. On Places365, where CaBM and V2C-CBM results are unavailable,
CBX-Bench ranks CBM-Suite highest among the original three CBMs. These results
illustrate the intended benchmark use case: once the MLLM council is validated
against human preferences, it can assign comparable CBX-Bench scores to
additional CBMs under the same evaluation protocol. This enables an online
benchmark in which future submissions are evaluated on the fixed image set and
ranked by per-dataset and overall explanation-quality scores.

Table~\ref{tab:benchmark-scores-scoring} also compares CBX-Bench with the
classification accuracy reported for each CBM. The rank order induced by CBX-Bench differs substantially from that induced by accuracy, indicating that CBX-Bench is not merely an accuracy proxy. For example, V2C-CBM achieves the highest reported ImageNet-100 accuracy but the lowest CBX-Bench score, whereas CaBM ranks first under CBX-Bench on both CUB-200 and ImageNet-100 despite ranking third by accuracy on both datasets.

\paragraph{Pairwise MLLM prompting is unsuitable as the main benchmark protocol.}
\begin{table}[t]
\centering
\caption{Position-bias instability under direct pairwise MLLM prompting.
Disagreement is the percentage of comparisons for which the preferred CBM label
changes after swapping the positions of the two explanations in the prompt.}
\label{tab:pairwise-position-bias}
\small
\begin{tabular}{lr}
\toprule
Judge & Disagreement (\%) \\
\midrule
Qwen3.5-9B & 17.2 \\
Qwen3.5-4B & 17.7 \\
Gemma4-E4B & 24.8 \\
InternVL3.5-8B & 29.8 \\
Gemma4-E2B & 33.2 \\
\bottomrule
\end{tabular}
\end{table}

We also test direct pairwise MLLM prompting, matching the human-study interface
with options for CBM A, CBM B, \emph{equally good}, and \emph{equally bad}.
Although this format is natural for human annotation, it is less suitable as the
main automated evaluation protocol for CBX-Bench. First, pairwise MLLM judges are
sensitive to candidate order~\cite{shi2025judging,wang2024large}. In our setting,
swapping the positions of the two CBM explanations changes the CBM-level MLLM
answer in 17.2\% to 33.2\% of comparisons, depending on the judge
(Table~\ref{tab:pairwise-position-bias}). Second, pairwise prompting produces
only a relative judgment between the two displayed explanations, rather than an
independent, fixed score for each CBM explanation. This makes benchmark rankings
dependent on the particular opponent set and pairings, and complicates the
evaluation of future CBM submissions under a common scoring protocol. We
therefore treat direct pairwise MLLM prompting as an ablation and use independent
score-based judging as the main CBX-Bench evaluation protocol.

\paragraph{Summary.}
The human study validates pairwise comparison as the reference protocol, while the MLLM study shows that independent score-based judging is the more stable automation layer. The 1--9 score-sum council recovers 70.7\% of strict human CBM rankings and 56.9\% of weak rankings that include ties, both substantially above the corresponding chance levels. This validated MLLM council then produces dataset-level CBM explanation benchmark rankings, while the pairwise prompting diagnostic explains why direct MLLM pairwise judging is not used as the main evaluation method.

\section{Discussion}
\label{sec:discussion}

Our results separate two questions that are often conflated in CBM evaluation: whether a model is accurate, and whether its displayed concepts form a useful explanation of its decision. Named concepts alone do not answer the second question. Their explanatory value depends on whether they are visually grounded, class-relevant, and convincing for the particular prediction being explained.

Pairwise preference makes this evaluation problem measurable without requiring ground-truth concept activations or a shared concept vocabulary. It also gives equality and disagreement a formal place in the protocol. This matters because many comparisons are genuinely close: some items lack annotator consensus, while others reach consensus that the two explanations are equivalent rather than that one CBM is better. Treating these outcomes as part of the signal avoids overstating the precision of explanation-quality claims.

The MLLM results show that this protocol can scale beyond direct human annotation, but only when the automated judge is designed carefully. Direct pairwise MLLM prompting is too sensitive to candidate order for our main benchmark. Independent scoring avoids this particular instability and lets us recover instance-wise rankings from the same human pairwise reference.

The resulting score-sum council rankings demonstrate the benchmark use case directly: the same validated protocol can compare CBM explanation quality across methods and datasets, rather than only reporting isolated qualitative examples.
\section{Limitations}
\label{sec:limitations}

Our human study covers three datasets, three CBMs, and 300 images. This is sufficient to expose consistent patterns in human and MLLM judgments, but it is not exhaustive over visual domains, concept vocabularies, or user populations.

The protocol measures relative preference, not absolute explanation sufficiency. A preferred explanation may still be incomplete, and a tie may mean that both explanations are useful or that neither is convincing. The rankings should therefore be read as comparative scores under the displayed evaluation interface, not as proof that the top-ranked explanation is fully faithful.

Finally, the MLLM findings are specific to the tested judges, scoring prompts, and aggregation rules. Stronger MLLMs may provide better score calibration, or separately validated judge selection may improve future benchmark reliability.

\section{Conclusion}
\label{sec:conclusion}

This work argues that CBM interpretability should be evaluated directly, rather
than inferred from predictive performance or illustrated through isolated
examples. Although CBMs expose human-understandable concepts, the presence of a
concept layer alone does not guarantee that the resulting explanations are
visually grounded, class-relevant, or helpful for understanding individual
predictions.

To address this evaluation gap, we developed \textbf{CBX-Bench} as a human-preference validated
benchmark for measuring CBM explanation quality. Our protocol uses pairwise human
preferences to define an interpretable reference signal, then uses this signal
to assess and select a score-based MLLM council for scalable evaluation. The
validated council assigns normalized explanation-quality scores, allowing CBMs
with different architectures and concept vocabularies to be compared under a
common protocol.

CBX-Bench therefore turns CBM explanation quality into a measurable benchmark
target. Instead of treating explanation quality as a qualitative supplement to
accuracy, it provides a fixed and auditable scoring procedure that can support
dataset-level rankings of existing methods and fair evaluation of future CBM
submissions.

\section{Acknowledgements}
We thank Ahmet Büyükyılmaz for his support in developing the human-study website. We also gratefully acknowledge the computational resources provided by the METU Center for Robotics and Artificial Intelligence (METU-ROMER).

\bibliographystyle{splncs04}
\bibliography{references}

\clearpage
\appendix
\section*{Appendix}
\renewcommand{\theHsection}{appendix.\Alph{section}}
\section{User Study Interface}
\label{app:user-study-ui}

\begin{figure}[h]
\centering
\includegraphics[width=.99\linewidth]{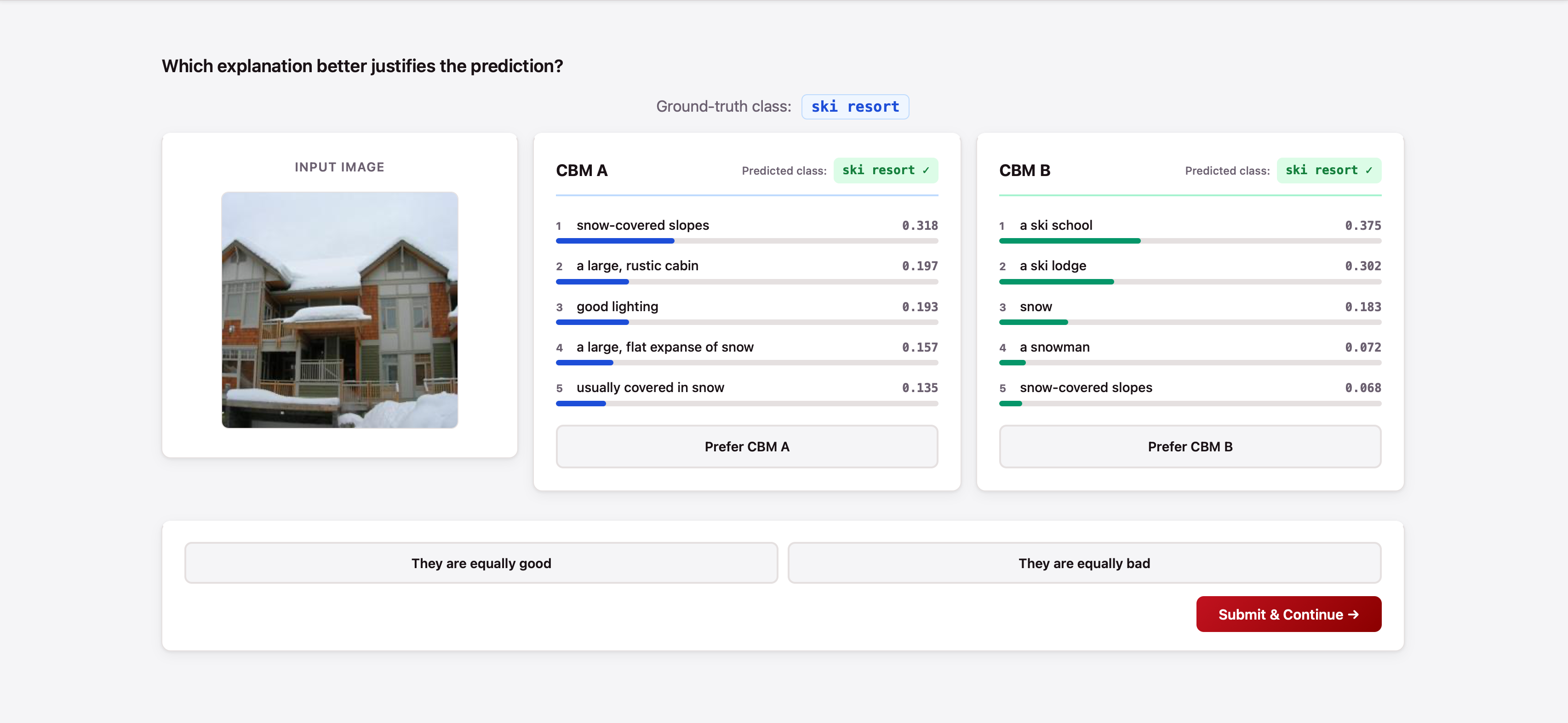}
\caption{User-study annotation interface for pairwise CBM explanation evaluation. Annotators see the input image, the ground-truth class, each CBM’s predicted class, and the top contributing concepts for each explanation. They then choose the more useful explanation, or mark the two explanations as equally good or equally bad.}
\label{fig:user-study-ui}
\end{figure}

\section{MLLM Judge Prompts}
\label{app:vlm-prompts}

This appendix reports the prompt templates used for MLLM judging. The image is provided as a multimodal input together with the text prompt. Braced fields denote item-specific values filled from the evaluation record. Independent score-based judging is the main protocol; the pairwise prompt is retained only for the order-sensitivity diagnostic reported in the results.

\subsection{Dataset Contexts}
\label{app:dataset-contexts}

The \texttt{[Dataset Context]} block in both judging prompts is filled with one of the following dataset-specific strings, depending on the evaluation subset:

\begin{PromptVerbatim}
CUB-200:
CUB-200-2011 bird image (200 fine-grained bird species)

ImageNet-100:
ImageNet natural image (1000 object categories)

Places365:
Places365 scene image (365 indoor and outdoor scene categories)
\end{PromptVerbatim}

\subsection{Independent Score-Based Judging}
\label{app:prompt-score-based}

For score-based judging, each CBM explanation is evaluated independently. We use the same template with 1--5, 1--7, and 1--9 integer rating scales. Pairwise labels are derived after inference by comparing the two independent scores. The prompt below follows the scoring prompt builder used in our experiments; only the system role, rating-scale block, and output score range change across scales.

The scale-specific system roles are:

\begin{PromptVerbatim}
1--5 scale:
You are an expert evaluator specialising in the interpretability of image
classification models. Your task is to rate the quality of a Concept Bottleneck
Model (CBM) explanation for a single input image on an integer scale from
1 (poor) to 5 (excellent).

1--7 scale:
You are an expert evaluator specialising in the interpretability of image
classification models. Your task is to rate the quality of a Concept Bottleneck
Model (CBM) explanation for a single input image on an integer scale from
1 (poor) to 7 (excellent).

1--9 scale:
You are an expert evaluator specialising in the interpretability of image
classification models. Your task is to rate the quality of a Concept Bottleneck
Model (CBM) explanation for a single input image on an integer scale from
1 (poor) to 9 (excellent).
\end{PromptVerbatim}

The remaining prompt body is shared:

\begin{PromptVerbatim}
[Task Description]
Concept Bottleneck Models (CBMs) decompose their predictions into
human-interpretable concepts. Each model activates semantic concepts from the
image and uses weighted concept activations to predict a class. You will see
the input image, the model's predicted class, and its top concept activations.

[Dataset Context]
{dataset description}

[Input Image]
The image to be classified is attached above.
Ground-truth class: {ground-truth class}

[CBM Explanation]
Predicted class : {CBM predicted class}
Concepts for the predicted class (top-5 activations L1-normalized to sum to 1,
then contributions < 0.010 removed; name -> normalized weight):
  1. {concept 1}  {normalized weight 1}
  2. {concept 2}  {normalized weight 2}
  ...
  5. {concept 5}  {normalized weight 5}

[Rating Scale]
{scale-specific rating block below}

[Evaluation Criteria]
Consider:

  - Concept Relevance   - Are activated concepts visually and semantically
    relevant to the predicted class?
  - Concept Faithfulness - Do weights reflect plausible discriminative
    importance?
  - Concept Diversity   - Do top concepts cover complementary aspects rather
    than redundancy?
  - Prediction Support  - Does the explanation justify the predicted class
    (whether correct or incorrect)?

[Output Format]
Respond with a single JSON object -- no prose before or after it:
{
    "reasoning": "<A concise explanation (1-2 sentences) for your rating>",
    "score": <integer from 1 to {maximum score}>
}
\end{PromptVerbatim}

The scale-specific rating blocks are:

\begin{PromptVerbatim}
1--5 scale:
  1 - Very poor: concepts are largely irrelevant or absent; explanation does
      not support the prediction.
  2 - Poor: mostly weak or misleading concepts; rationale is unconvincing.
  3 - Adequate: some relevant concepts but incomplete, redundant, or weakly
      weighted.
  4 - Good: relevant, diverse concepts with reasonable weights that support
      the prediction.
  5 - Excellent: highly faithful, coherent explanation a domain expert would
      find convincing.

1--7 scale:
  1 - Extremely poor: concepts are mostly irrelevant, absent, or misleading;
      the explanation does not support the prediction.
  2 - Very poor: only minimal useful evidence; misleading or irrelevant
      concepts dominate.
  3 - Poor: a few concepts may be relevant, but the explanation is weak,
      incomplete, or poorly weighted.
  4 - Adequate: contains a reasonable mix of relevant concepts, but with
      noticeable gaps, redundancy, or imperfect weighting.
  5 - Good: mostly relevant and coherent, with plausible weights that support
      the prediction.
  6 - Very good: strong, faithful, and coherent explanation with only small
      imperfections.
  7 - Excellent: highly faithful, coherent explanation a domain expert would
      find convincing.

1--9 scale:
  1 - Extremely poor: concepts are mostly irrelevant, absent, or misleading;
      the explanation does not support the prediction.
  2 - Very poor: only minimal useful evidence; misleading or irrelevant
      concepts dominate.
  3 - Poor: a few concepts may be relevant, but the explanation is weak,
      incomplete, or poorly weighted.
  4 - Somewhat weak: partially supports the prediction, but important visual
      evidence is missing or concept weights are questionable.
  5 - Adequate: contains a reasonable mix of relevant concepts, but with
      noticeable gaps, redundancy, or imperfect weighting.
  6 - Fairly good: mostly relevant and coherent, with minor omissions or
      some less useful concepts.
  7 - Good: relevant, diverse concepts with plausible weights that support
      the prediction well.
  8 - Very good: strong, faithful, and coherent explanation with only small
      imperfections.
  9 - Excellent: highly faithful, coherent explanation a domain expert would
      find convincing.
\end{PromptVerbatim}

The output score range is instantiated as:

\begin{PromptVerbatim}
1--5 scale:
{
    "reasoning": "<A concise explanation (1-2 sentences) for your rating>",
    "score": <integer from 1 to 5>
}

1--7 scale:
{
    "reasoning": "<A concise explanation (1-2 sentences) for your rating>",
    "score": <integer from 1 to 7>
}

1--9 scale:
{
    "reasoning": "<A concise explanation (1-2 sentences) for your rating>",
    "score": <integer from 1 to 9>
}
\end{PromptVerbatim}

\subsection{Pairwise Judging}
\label{app:prompt-pairwise}

\begin{PromptVerbatim}
[System Role]
You are an expert evaluator specialising in the interpretability of image
classification models. Your task is to compare the explanations produced by
two Concept Bottleneck Models (CBMs) for the same input image and determine
which model provides a more faithful, coherent, and human-understandable
explanation of its prediction.

[Task Description]
Concept Bottleneck Models (CBMs) decompose their predictions into
human-interpretable concepts. Each model first activates a set of semantic
concepts from the image and then uses those concept activations (weighted by
learned importance scores) to make a final class prediction. A high-quality
CBM explanation should:
  (1) Activate concepts that are visually present and semantically relevant
      to the predicted class;
  (2) Assign higher weights to concepts that are most discriminative for
      the prediction;
  (3) Collectively form a coherent, self-consistent rationale that a domain
      expert would find convincing.

[Dataset Context]
{dataset description}

[Input Image]
The image to be classified is attached above.
Ground-truth class: {ground-truth class}

[CBM1 Explanation]
Predicted class : {CBM 1 predicted class}
Concepts for the predicted class (top-5 activations L1-normalized to sum to 1,
then contributions < 0.010 removed; name -> normalized weight):
  1. {CBM 1 concept 1}  {CBM 1 normalized contribution 1}
  2. {CBM 1 concept 2}  {CBM 1 normalized contribution 2}
  ...
  5. {CBM 1 concept 5}  {CBM 1 normalized contribution 5}

[CBM2 Explanation]
Predicted class : {CBM 2 predicted class}
Concepts for the predicted class (top-5 activations L1-normalized to sum to 1,
then contributions < 0.010 removed; name -> normalized weight):
  1. {CBM 2 concept 1}  {CBM 2 normalized contribution 1}
  2. {CBM 2 concept 2}  {CBM 2 normalized contribution 2}
  ...
  5. {CBM 2 concept 5}  {CBM 2 normalized contribution 5}

[Evaluation Criteria]
When comparing the two explanations, consider the following criteria:

  - Concept Relevance   - Are the activated concepts visually and semantically
    relevant to the predicted class?
  - Concept Faithfulness - Do the concept weights reflect their true
    discriminative importance (i.e., would a human expert assign similar
    importance to these concepts)?
  - Concept Diversity   - Do the top concepts together cover distinct and
    complementary aspects of the predicted class, rather than being redundant?
  - Prediction Support  - For a correct prediction, does the explanation
    adequately justify why the model chose that class? For an incorrect
    prediction, does the explanation reveal the conceptual confusion?

[Decision Rules]
Choose exactly one label:

- CBM1: CBM1's explanation is meaningfully better than CBM2's explanation.
- CBM2: CBM2's explanation is meaningfully better than CBM1's explanation.
- Equally_Good: both explanations are visually grounded, coherent, and similarly
  useful for understanding the model prediction.
- Equally_Bad: both explanations are weak, hallucinated, irrelevant, redundant,
  or fail to support or diagnose the model prediction.

A small local advantage is not enough to choose CBM1 or CBM2. Prefer a CBM only
when the quality gap would matter to a careful domain expert.

When the two explanations are close in quality, first decide whether both are
acceptable or both are problematic:
- if both would be acceptable explanations, choose Equally_Good;
- if both would be rejected or treated with serious caution, choose Equally_Bad.

Important: do not use Equally_Good as a generic "no clear winner" label. If
neither explanation is clearly good, use Equally_Bad.

[Output Format]
Respond with a single JSON object -- no prose before or after it:
{
    "reasoning": "<A concise explanation (1-2 sentences) of your comparative assessment of the two explanations>",
    "preferred_cbm": "<Exactly one of: CBM1, CBM2, Equally_Good, Equally_Bad>"
}
\end{PromptVerbatim}

\end{document}